\documentclass[letterpaper, 10 pt, conference]{ieeeconf}  
\IEEEoverridecommandlockouts 
\usepackage{times} 
\usepackage{amsmath} 
\usepackage{amssymb}  
\usepackage{outlines}
\usepackage{times}
\usepackage{url}
\usepackage{graphicx}
\usepackage{float}
\usepackage{makebox}
\usepackage{algorithm}
\usepackage{algorithmic}

\newcommand{\piPLAN}{\pi_{\textsc{PLAN}}}
\newcommand{\piMF}{\pi_{\textsc{SKILL}}}
\title{\LARGE \bf
Learning Skills to Patch Plans Based on Inaccurate Models
}

\author{Alex LaGrassa$^{1}$, Steven Lee$^{1}$, and Oliver Kroemer$^1$ \thanks{$^{1}$Robotics Institute, School of Computer Science, Carnegie Mellon University, Pittsburgh, PA, USA {\tt\small \{alagrass, stevenl3, okroemer\}@cs.cmu.edu}}
}%

\begin{document}

\maketitle
\thispagestyle{empty}
\pagestyle{empty}

%

\begin{abstract}
Planners using accurate models can be effective for accomplishing manipulation tasks in the real world, but are typically highly specialized and require significant fine-tuning to be reliable. Meanwhile, learning is useful for adaptation, but can require a substantial amount of data collection. In this paper, we propose a method that improves the efficiency of sub-optimal planners with approximate but simple and fast models by switching to a model-free policy when unexpected transitions are observed. Unlike previous work, our method specifically addresses when the planner fails due to transition model error by patching with a local policy only where needed. First, we use a sub-optimal model-based planner to perform a task until model failure is detected. Next, we learn a local model-free policy from expert demonstrations to complete the task in regions where the model failed. To show the efficacy of our method, we perform experiments with a shape insertion puzzle and compare our results to both pure planning and imitation learning approaches. We then apply our method to a door opening task. Our experiments demonstrate that our patch-enhanced planner performs more reliably than pure planning and with lower overall sample complexity than pure imitation learning.
\end{abstract}

\section{Introduction}
The ability for robots to adapt to changing needs and conditions in human environments is necessary for expanding their utility into new application domains. A robot can be pre-programmed with general models and reasoning capabilities before deployment, but some amount of adaptation is necessary to capture the wide range of conditions a robot may encounter. 

Motion planners with accurate models are often used to accomplish tasks, but are often highly specialized and require significant fine-tuning to be reliable \cite{kavraki2016motion, mericcli2012experience}. Inaccuracies or deviations in a system’s model can increase the complexity of the controller, and the potential for failed task executions. Contact-rich manipulation tasks are difficult to model because of the intricacies of changing contact modes \cite{ji2001planning}. Nonetheless, planners are still useful when the robot maintains contact, as modeling some phenomena, such as friction on a sliding surface, can be sufficiently accurate to provide the planner with useful information \cite{goyal1991planar, chavan2020planar}. However, more complex interactions may lead to the model, and hence the planner, failing at execution time.

Learning-based approaches allow robots to adjust executions through experience, whether through supervised demonstrations or reinforcement learning. However, the amount of data required to acquire learned skills that generalize can be both large and often non-trivial to collect~\cite{finn2017one, lee2019making, rajeswaran2017learning}.

In this work, we propose a framework that facilitates motion planning with sub-optimal planners, limited training data, and inaccurate system models. Our method combines model-based planning with model-free learning by first identifying states where the transition model provides a poor estimate and then learning a local policy, e.g., a skill, to patch the task execution. The robot can then reliably perform the task by predicting if the original plan will lead to model failure, then adapt the plan and patch it with a learned skill if needed. Learning skills that only need to generalize across a small part of the state space can require fewer samples than a more general learned skill would need.

Integrating planning and learning in this manner enables the robot to acquire new skills only when needed. The new skill is data-efficient to learn as it only generalizes across the relatively narrow range of conditions where the model fails. This hybrid approach maintains the benefit from the broad generalization afforded by the model-based planning in regions where the model is accurate.

We evaluate our method on a shape insertion task and a door opening task, comparing our results to the performance of pure planning and imitation learning approaches.

\begin{figure}[t]
\centering
\includegraphics[width=0.45\textwidth]{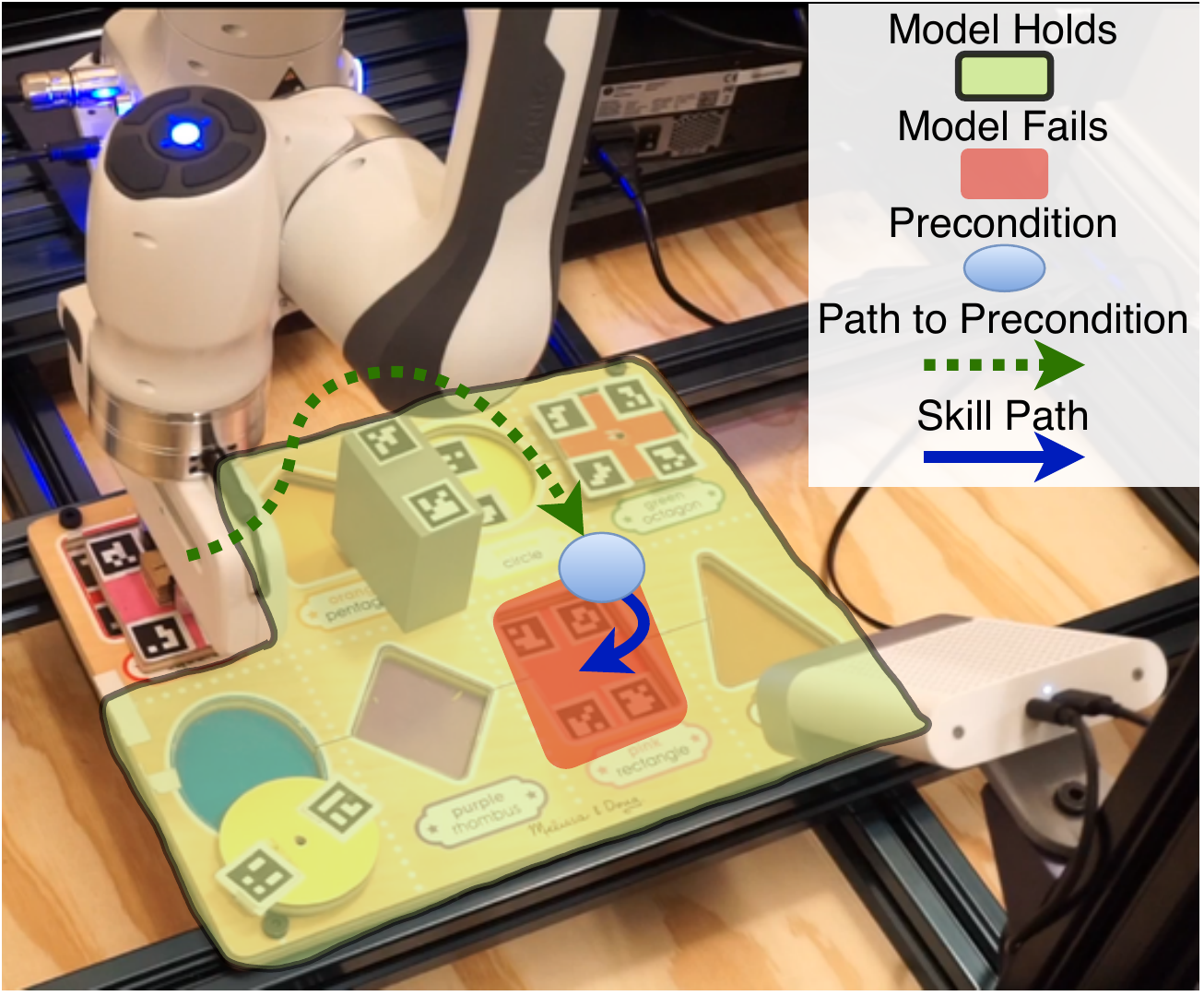}
\caption{ Our method follows a trajectory to a precondition and then executes a learned skill to complete the shape insertion task in the region where the system model fails. A plastic gray box obstructs the path to the goal. An overhead (not shown in image) and angled camera are aimed at the center of the board.}
\label{fig:envsetup}
\vspace{-1.5em}
\end{figure}

\section{Related Work}
\label{sec:RelatedWork}
\textbf{Combining Model-Free \& Model-Based Approaches:} The most closely related work in combining model-based planning with model-free learning to ours is Lee et al.'s Guided Uncertainty-Aware Policy Optimization~\cite{guapo}. Lee et al.\ use model-based planning to reach an uncertain region, which is defined by uncertainty in the observation model, and then switch to a model-free policy for a real-robot peg insertion task. Lee's work focuses on perception uncertainty, while our work directly addresses planning failure due to transition model errors. Other approaches use model-based planning as an exploration policy to more efficiently collect meaningful data samples that can be used for learning a model-free policy~\cite{shyam2018model, rajeswaran2017learning, NIPS2014_5444}. Adaptive Online Planning also combines model-based planning with model-free learning for a highly accurate, but computationally expensive, planner that can obtain trajectories~\cite{lu2019adaptive}. In contrast, our approach is intended for planners that have have approximate models but are relatively inexpensive to query. Hoppe et al.\ use active learning as part of the trajectory optimization so the planner can select more informative samples for model-free policies~\cite{hoppe2019planning}.

\textbf{Policy Composition:} Combining and chaining policies is commonly formalized using the \textit{options} framework~\cite{sutton1999between}. Each option has an associated policy, precondition, and set of termination conditions. Konidaris et al.\ chain policies to form more complex behaviours and examine \textit{option discovery}~\cite{konidaris2009skill}. Option discovery is a problem where multiple options are learned and added to a system where needed. Most importantly, options can be combined hierarchically to execute complex behaviors~\cite{nachum2018data}.

\textbf{Anomaly Detection:} Our method uses anomaly detection to determine when a model deviation has occurred during execution. Multimodal monitoring has been shown to be more reliable than single mode monitoring of unexpected observations~\cite{wu2019analysis}. Park et al.\ use this concept as a flag to terminate executions after an anomaly has been detected~\cite{park2018multimodal}. Vemula et al. use a history of where the model has failed to avoid those states while planning~\cite{vemula2020planning}. In contrast to previous works, our approach uses anomaly detection with a model-based planning method to determine which states require a local policy using model-free learning.

\textbf{Exploiting Contacts for Manipulation Tasks:} Guan et el.\ leverage contacts to reduce the number of states considered during planning to scale with state complexity. Guan et al.\ also model the problem as a composite MDP (Markov Decision Process) in \textit{SE}(2) with added domain-specific structures to enable a solution using dynamic programming~\cite{guan2018efficient}. Many approaches use contact modes as a variable during optimization for complex contact-rich locomotion and manipulation tasks~\cite{toussaint2018differentiable,posa2014direct, mordatch2012discovery}. 
However, the resulting trajectories can be difficult to execute reliably in the presence of modeling errors. P\'all et al.\ use the Contingent, Contact-Exploiting RRT (ConCERRT) framework to find a path that accounts for uncertainty in the transition model, attempting to find a contingency plan for every possible belief state~\cite{pall2018contingent}.

\textbf{Model-Free Skill Learning:} Contact rich manipulation has also seen advances through the use of model-free reinforcement learning \cite{rajeswaran2017learning, andrychowicz2020learning, lee2019making}. However, deep reinforcement learning often requires large amounts of training data. Imposing structure on problems can help improve sample efficiency when performing model-free optimization. Constraint Optimization and Reinforcement Learning (CORL) employs a user-specified low-dimensional projection that provides structure to make reinforcement learning more efficient \cite{englert2016combined}. Englert et al.\ later use CORL to train a cabinet opening skill from a single demonstration \cite{englert2018}. Additionally, sample efficiency may be improved through defining the policy using fewer parameters \cite{da2012learning}. Our method describes how these advances in model-free learning can be integrated into systems that plan using models. 
\begin{figure}[t]
\vspace{0.8em}
\centering
\includegraphics[width=0.48\textwidth]{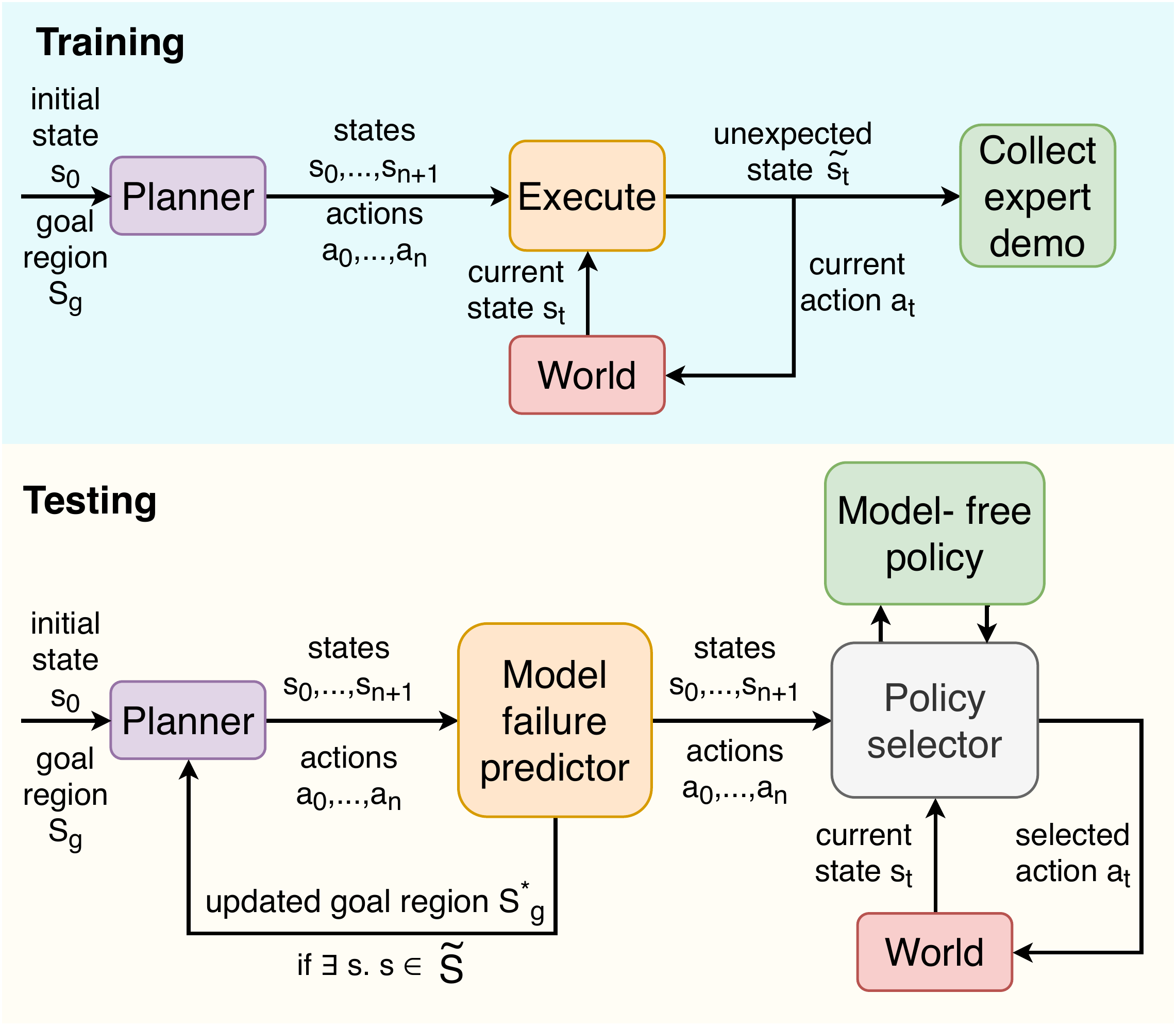}
\caption{System block diagram. We use the training phase to collect data and fit models for use in the testing phase. The planner gives a sequence of expected states and actions to the execution module, which checks for anomalous state observations and requests expert demonstrations after such observations occur. We refer to an unexpected state at time t as $\Tilde{s}_t$ in this diagram for clarity.}
\label{fig:overall}
\vspace{-2em}
\end{figure}

\section{Problem Statement}
\label{sec:ProblemStatement}
We formulate the problem as a planning problem with a state space $\mathcal{S}$, an action space $\mathcal{A}$, and a starting state $s_0 \in \mathcal{S}$. The robot must compute a finite list of actions $[a_0, a_1, \ldots a_n]$ executed in states $[s_0, s_1, \ldots s_n]$, such that the final state, $s_{n+1}$, is in a goal set, $\mathcal{S}_g \subseteq \mathcal{S}$. $\piPLAN$ maps a state, $s_t$, deterministically to the next action, $a_t$, according to the plan. We have access to an approximate transition model, but not the underlying dynamics. The transition model we use is $\widehat{\mathcal{T}} : \mathcal{S} \times \mathcal{A} \rightarrow \mathcal{S}$, which is an estimate of $P(s_{t+1}|s_{t},a_{t})$ and is not expected to exactly follow the real world distribution.
Using $\widehat{\mathcal{T}}$ and $s_0$, the planner conducts a search to a set of goal states, $\mathcal{S}_{g}^*$. $\mathcal{S}_{g}^*$ is an intermediate goal. If the planner is being used to solve the planning problem without the model-free policy, then $\mathcal{S}_{g}^* = \mathcal{S}_g$. 

The planner can generate a suitable plan if the transition model is sufficiently accurate. If the model fails during the execution then the robot should stop following the plan and use a different strategy. Therefore, our approach determines for which $\widetilde{\mathcal{S}} \subseteq \mathcal{S}$ the agent should stop following $\piPLAN$ and instead learn (during training) or execute (during testing) the model-free policy, $\piMF$. $\piMF$ is learned from a series of expert actions, $[\hat{a}_t, \hat{a}_{t+1}, \ldots, \hat{a}_{m}], [\hat{s}_t, \hat{s}_{t+1}, \ldots, \hat{s}_{m+1}]$. In this work, we sometimes refer to these policies as a \textit{skill}, but symbolically represent them as $\piMF$ to indicate that the skill is a learned policy. 

\vspace{-0.5em}

\section{Technical Approach}
\vspace{-0.5em}
\label{sec:approach}

\begin{algorithm}[t]
\caption{Training procedure}
    \begin{algorithmic}
        \STATE $\mathcal{D}_s \gets \{\}$
        \STATE $\widetilde{\mathcal{D}}_s \gets \{\}$
        \STATE $\mathcal{S}^*_g \gets \mathcal{S}_g$ 
        \STATE precompute $\piPLAN(\mathcal{S}^*_g)$
        \FOR{$t=0$ to $N$}
                \STATE $s_{t+1} \gets \textsc{EXECUTE}(a_t)$
                \IF{$P(s_{t+1}\,|\,s_t, a_t) > p$} 
                    \STATE $\mathcal{D}_{s} \gets \mathcal{D}_{s} \cup \{s_t\} $
                \ELSE
                    \STATE $\widetilde{\mathcal{D}_{s}} \gets \widetilde{\mathcal{D}_{s}} \cup \{s_t\} $
                    \STATE $\mathcal{D}_{\mathcal{I}} \gets \mathcal{D}_{\mathcal{I}} \cup  (\hat{s}_t,\hat{a}_t)$ from expert demonstration
                    \STATE \textbf{break}
                \ENDIF
            \ENDFOR
    \STATE $\piMF \gets \textsc{TRAIN}(\mathcal{D}_{\mathcal{I}})$

    \end{algorithmic}
\label{fig:pseudocodetraining}
\end{algorithm}

\subsection{Overall Approach}
\label{sec:overall}
Our method is designed to reliably complete a robotic manipulation task using sub-optimal models and limited real-world robot data. A model-based planner is used until model failure is detected and then a policy is learned from demonstrations to reach the goal. At test time, the robot executes the learned skill in the regions where the model has failed. Thus, much of the task is initially completed using the model-based planner to increase generalization and decrease data needs. If no model failure occurs, then no skill is learned. 
However, if the robot encounters or predicts model failure, a skill is learned to patch the plan and complete the task.

At training time, state regions are gathered where the transition model $\widehat{\mathcal{T}}$ holds, $\mathcal{D}_s$, and does not hold, $\widetilde{\mathcal{D}_{s}}$. While training, we use $\piPLAN$ to obtain samples of $(s_{t+1}, s_t, a_t)$. These samples are labeled according to a confidence interval, $P(s_{t+1} \,|\, s_t, a_t) > p$, where $p$ is set to 0.98 for our experiments. Our transition model for action $a_t$ from $s_t$ over $s_{t+1}$ is modeled as $\mathcal{N}(s_{t+1}, k_0|s_{t+1}-s_{t}|)$. This allows the acceptable error to grow proportionally with the distance of the movement, where $k_0$ is a proportionality constant that can be determined experimentally by fitting expected deviation from the target $s_t$ for different state transitions. We show pseudocode for training our method in Algorithm~\ref{fig:pseudocodetraining}.

The action and state spaces for anomaly detection are high-level; anomaly detection is executed only after the joint trajectory controller has terminated. We use a joint impedance controller so if the controller deviates from the expected path between $t$ and $t+1$, but corrects itself before the next anomaly detection step, then $s_{t+1}$ is not in $\widetilde{\mathcal{S}}$.
If a sample is within the confidence interval, then the corresponding state is added to $\mathcal{D}_s$. Otherwise, the sample is added to $\widetilde{\mathcal{D}_{s}}$. $\widetilde{\mathcal{D}_{s}}$ and $\mathcal{D}_{s}$ are then used to estimate $\widetilde{\mathcal{S}}$ using a Gaussian Process~\cite{williams2006gaussian}.

When a state is added to $\widetilde{\mathcal{D}_{s}}$, the model has failed, and the human operator is asked to provide a training demonstration for completing the task from this state. The skill demonstration is added to the imitation learning dataset, $\mathcal{D}_{\mathcal{I}}$. This dataset is used to estimate the skill $\piMF$ and its corresponding initiation set, $\cal{I}_{\textsc{SKILL}}$, a distribution over starting states which the learned skill is likely to achieve the goal from. We further elaborate on the skill learning in Section~\ref{sec:modelfree}. 

At test time, $\piPLAN$ outputs a series of states. If any states are estimated to be in $\widetilde{\mathcal{S}}$ then $\piPLAN$ instead computes a path to $S^*_g$, which is now $\cal{I}_{\textsc{SKILL}}$. 
In our experiments, the expert demonstrations for shape insertion produced better results when the shape was reset to a position outside of the hole. The shape was then inserted into the hole using a sliding motion, as shown in Fig.~\ref{fig:demosetup}. This insight motivated our decision to have the agent change $\widetilde{\mathcal{S}}_{g}^*$ to the initiation set.
Our test algorithm is shown in Algorithm~\ref{fig:pseudocodetest}. In the following sections, we describe each step in more technical detail.

\begin{algorithm}[t]
\caption{Testing procedure}
    \begin{algorithmic}
        \STATE $\mathcal{S}^*_g \gets \mathcal{S}_g$ 
        \STATE precompute $\piPLAN(\mathcal{S}^*_g)$
        \FOR{$t=0$ to $N$}
                \IF{$s_t \in \widetilde{\mathcal{S}}$} 
                    \STATE $\mathcal{S}_{g}^* \gets \cal{I}_{\textsc{skill}}$
                    \STATE \textbf{break}
                \ENDIF
        \ENDFOR
        \STATE $t \gets 0$
        \WHILE{$s_t \notin \mathcal{S}_{g}^*$ and $s_t \in \widetilde{\mathcal{S}}$ or $s_t \in \mathcal{I}_{\textsc{skill}}$}
                \STATE $s_{t+1} \gets \textsc{EXECUTE}(a_t)$
                \STATE $t \gets t+1$
        \ENDWHILE
        \WHILE{$s_t \notin \mathcal{S}_{g}^*$}
            \STATE $a'_t \sim \piMF(s_t)$
            \STATE $s_{t+1} \gets \textsc{EXECUTE}(a'_t)$ 
            \STATE $t \gets t+1$
        \ENDWHILE
    \end{algorithmic}
\label{fig:pseudocodetest}
\end{algorithm}

\vspace{-2em}

\begin{figure}[b]
\vspace{-1em}
\centering
\includegraphics[width=0.45\textwidth]{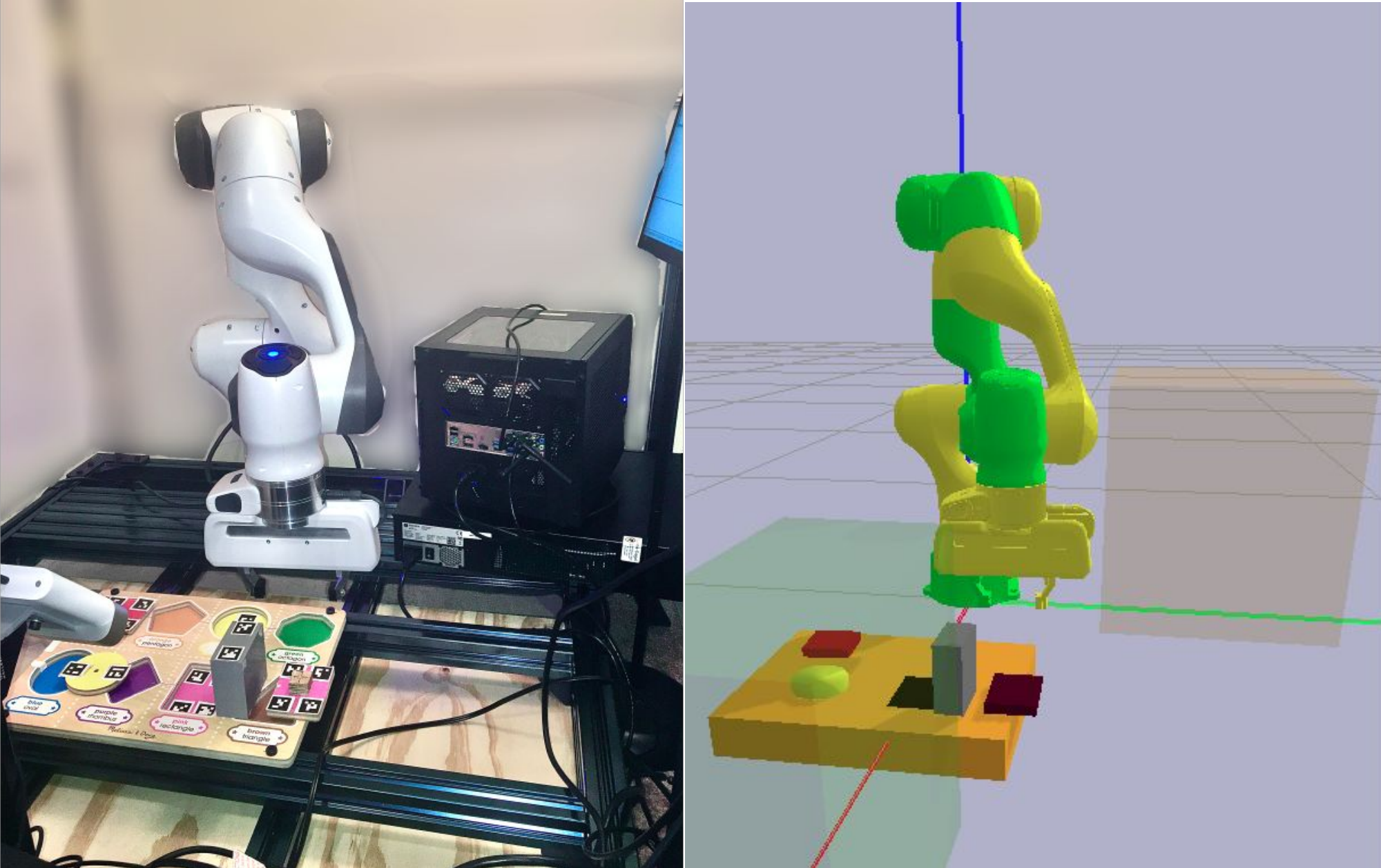}
\caption{The real world environment (left) and the corresponding Bullet3D environment for planning and collision detection (right).}
\label{fig:pybullet}
\end{figure}

\subsection{Model-based Planner}
\label{sec:model-based-planner}
The purpose of the model-based planner is to quickly find a sequence of waypoints, $s_0, s_1, \ldots, s_{n+1}$ and corresponding $n$ actions to get from each $s_t$ to $s_{t+1}$ such that $s_{n+1} \in \mathcal{S}^*_g$  according to $\widehat{\mathcal{T}}$. 

Planning is done using a PyBullet simulation~\cite{coumans2019} to detect collisions and predict transitions, as shown in Fig.~\ref{fig:pybullet}. 
The planner has access to approximate CAD models of the objects and their poses. We also assume that objects do not deform during planning.

The high level planner samples grasps and inverse kinematics solutions, and then chooses the grasp that minimizes $|q_0 - q_{n+1}|$, the distance between start and end joint configurations, conditioned on a collision-free plan existing for the task. This high level sampling is important during highly constrained placement scenarios, such as the one shown in Fig.~\ref{fig:envsetup}, as not all grasps will have a final feasible plan. This long-horizon reasoning facilitates generalization across different situations even in constrained domains. 

The motion planner used to achieve goal states is a bi-directional RRT (Rapidly Exploring Random Trees)~\cite{rrt} in order to return a collision-free solution quickly and with high probability. We apply smoothing and restarts to improve the quality of the trajectories: 5 restarts, 100 smoothing iterations, and 200 search iterations for motion planning. We built our planning stack using library tools provided by Garrett ~\cite{pybulletplanining}.

For the door handle turning tasks, we specify that the trajectory is an interpolation in \textit{SE}(3) between the start handle pose and the end handle pose once grasped. If the initial configuration or end configuration is in collision, our planner returns $\{\}$. Additionally, if $s_{t+1}$ is expected to penetrate an object in simulation, then contact is also predicted. End configurations can be sampled if there is at least one goal state, i.e. $|\mathcal{S}_g| >1$.

\begin{figure}[t]
\includegraphics[width=0.48\textwidth]{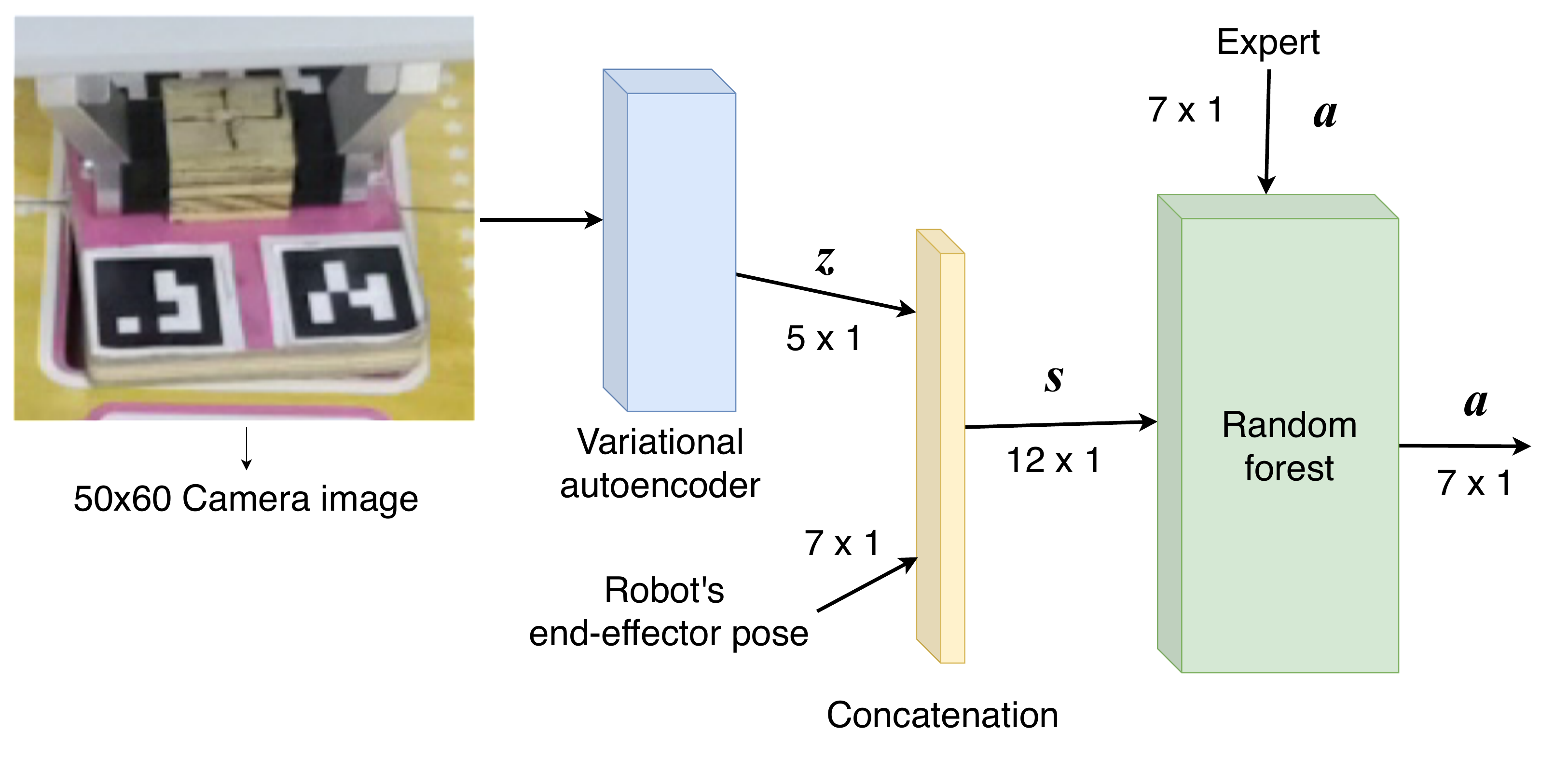}
\caption{Dataflow and architecture for model free policy.}
\label{fig:dataflow}
\vspace{-1em}
\end{figure}

\subsection{Model failure detection during execution}
\label{sec:ModelFailureDetection}
At the beginning of the plan execution, we initialize two datasets for classification: one dataset with expected $s_{t+1}$ given $(s_t, a_t) \sim \piPLAN$ ($\mathcal{D}_{\mathcal{S}}$) and one dataset with unexpected $s_{t+1}$ based on the transition model used by the planner, ($\mathcal{D}_{\mathcal{S}}$). 

We measure state using our multimodal perception system, which uses vision, joint state estimation, and contact forces. Coordinating across multiple modes addresses partial observability within individual sensor modalities. For instance, precisely measuring the 6 DOF pose of the manipulator is difficult when the robot is grasping it, but binary contact detection and estimation of the robot's end-effector pose is trivial. Thus, we use both deviation of the end-effector in Cartesian space and binary contact sensing for anomaly detection. We describe our perception system in detail in Section~\ref{sec:experimental}. 

At test time, if the agent predicts that $\exists s \in [s_0, s_1, \ldots, s_{n}]$ such that $s \in \widetilde{\mathcal{S}}$, then the planner replans to the skill's initiation set $\cal{I}_{\textsc{SKILL}}$ and then samples actions from the skill policy $\piMF$ until the task is complete.

We use Gaussian Process (GP) regression on a decision rule, $g(s_t)$, to predict whether $s_{t} \in \widetilde{\mathcal{S}}$ using a GP, $f$. The inputs $\mathbf{x}$ are rows of $[s_t]$. The labels, denoted as $\mathbf{y}$, are our decision rule, $g(s_t)$ where $g(s_t) = 1$ if $s_t 
\in \widetilde{\mathcal{D}_{\mathcal{S}}}$ and $0$ otherwise. 
The kernel is a 5/2 Mat\'{e}rn kernel, a special case for which computation is very efficient. The kernel function is shown in Eq.~\ref{eq:kernel} where $d$ is the distance between $x$ and  $x'$, such as $|x-x'|$~\cite{williams2006gaussian}. We optimize $\theta$, the hyperparameters of the kernel, which include $\sigma$ and $\rho$ for each dimension, using Large-scale Bound-constrained Optimization (L-BFGS-B)~\cite{zhu1997algorithm}. Equation~\ref{eq:mle} describes the likelihood to be maximized.

\begin{equation}
\label{eq:kernel}
K(d) = \sigma^2
       \left( 1 + \frac{\sqrt{5}d}{\rho}+\frac{5d^2}{3p^2} \right)
       \exp \left(-\frac{\sqrt{5}d}{\rho} \right)
\end{equation}

\begin{equation}
\label{eq:mle}
\log p(\mathbf{y}|\theta, \mathbf{x}) = -\frac{N \log 2\pi}{2} - \frac{\log \det(K+\sigma^2_nI)}{2} - \frac{\mathbf{y}^{T}K^{-1}\mathbf{y}}{2}
\end{equation}

The threshold for $g(x)$ to indicate model failure is an application-specific hyperparameter, $0 \leq \tau \leq 1$. We chose $\tau = 0.75$ to be conservative, because model failure in the shape insertion domain can lead to movement of the shape in the robot's gripper, making it less likely that $\piMF$ will succeed.


\subsection{Model-free Skill Learning}
\label{sec:modelfree}

Once the robot encounters a model failure during training, it notifies a human operator to help finish the task and collects samples of $(\hat{s}_t, \hat{a}_t)$, which are added to $\mathcal{D}_{\mathcal{I}}$. The human gives the robot keyboard teleoperated actions to go to a skill starting location and then complete the task. Uniform random noise between $[-\beta, +\beta]$ was added to each action executed. $\beta$ needs to be high enough to make the skill reliable, but not so high that the human cannot complete the task. 
We found that the policy produced from this demonstration data alone, without the noise, had a limited distribution of visited states, leading to poor generalization in unfamiliar states. This noise injected demonstration approach was inspired from DART (Disturbances for Augmenting Robot Trajectories) \cite{laskey2017dart}, which showed that noise injection lead to a more robust policy.

\begin{figure}[t]
\vspace{1em}
    \includegraphics[width=0.36\textwidth]{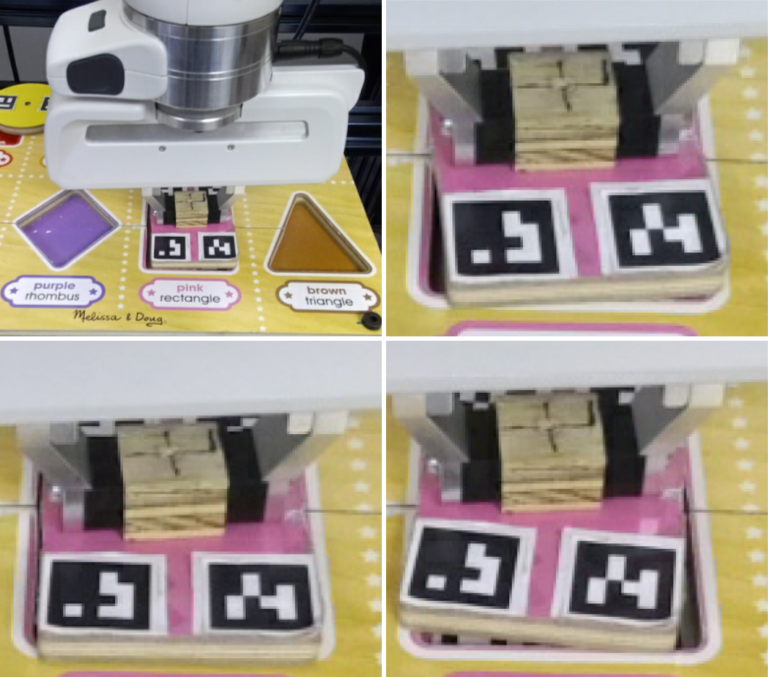}
    \centering
\vspace{-1em}
    
    \caption{Camera images corresponding to states where the transition model failed. These are used as inputs to the autoencoder mentioned in Section~\ref{sec:modelfree}. Not all variations of model failure are depicted.}
    \label{fig:failureexamples}
\vspace{-2.5em}
\end{figure}

For state representation, we use a variational autoencoder, which learns an embedding that can be used to reconstruct an image~\cite{kingma2013auto}. The inputs to the autoencoder are camera images of the scene taken as the demonstrations are being performed. Examples of these images are shown in Fig.~\ref{fig:failureexamples}. The embeddings produced, along with the end-effector positions, are then used as inputs to $\piMF$. We show the data flow in Fig.~\ref{fig:dataflow}. The autoencoder loss is the mean-squared error between the original image and the reconstructed image. The architecture is a convolutional layer with 32 $5\times5$ filters, 16 $3\times3$ filters, and a Gaussian noise layer with $\sigma=0.001$. The output is flattened, and then passed through two more fully connected layers, $\mu_z$ and $\sigma_z$. We use the re-parameterization trick from~\cite{kingma2013auto} to sample from a unit Gaussian, $\epsilon \sim \mathcal{N}(0,I)$ and then sample a latent vector $z$ as $\mu_z + \epsilon \sqrt{\sigma_z}$. We chose $z$ to be 5-dimensional. 
The $z$ was reconstructed to the original image size using fully connected layers. We based our architecture on~\cite{dsae}. Because our dataset is small, we also perform data augmentation from~\cite{scikit-learn} to reduce overfitting, including rotation, shear transforms, color, and random noise to increase our dataset size by a factor of 50.

After the data is collected, we use $\mathcal{D}_{\mathcal{I}}$ to fit a function $\piMF := \mathcal{S} \rightarrow A$ that maps each state to an action. $\piMF$ can also be represented using other policy forms, such as neural networks or Dynamic Movement Primitives~\cite{kroemer2016meta}. We chose random forest regression (RFR) for our model-free policy due to its interpretability and data efficiency~\cite{jia2019cloth}. The representational ability of RFR was sufficient for our tasks. 

RFR outputs numerical values instead of class labels, using the mean-squared generalization error. The random forest takes in training data, $\mathbf{X}$, and outputs a tree predictor, $h(x)$ that minimizes the mean-squared generalization error, $\mathbb{E}_{X,Y} \left[ \left(Y - h(X) \right)^2 \right]$ over all available demonstrations. Each decision tree in the random forest uses a randomly selected subset of the features. $\mathbf{X}$ is our $\dim(z) \times N$ latent visual features concatenated with the $\dim(a) \times N$ end-effector pose, to predict a $\dim(a) \times N$ action, which is our $\textbf{Y}$. We perform grid-search cross-validation for hyper-parameter optimization, which include the number of trees, maximum number of features for splitting, minimum number of samples to be at an internal node, minimum number of samples to be at a leaf node, and maximum tree depth. We use the RFR implementation in scikit-learn~\cite{scikit-learn}.

We fit the initiation set to be a Gaussian distribution: $\mathcal{N}(\overline{s_0}, \Sigma_{s_0})$. Initial starting states are sampled from  $\mathcal{I}_{\textsc{skill}}$ when setting $\mathcal{S}^*_g$ after model failure or expected model failure.
\vspace{-1em}
\section{Results}
In this section, we describe the experiments used to evaluate our proposed method of combining model-based planning and model-free skill learning.
For both tasks, the action is a 3D translation $\Delta x$ of the end-effector in the world frame and a rotation represented using a quaternion.

\subsection{Experimental Setup}
\label{sec:experimental}
\textbf{Shape Insertion Setup:} The experimental setup for the shape insertion experiment is shown in Figure~\ref{fig:envsetup}. We use a children's 8-piece knob puzzle (or \textit{board}) mounted at a fixed location. The puzzle pieces are manipulated by a 7 DoF Franka Panda arm towards their respective goal locations. A 7cm x 8cm x 4cm PLA obstacle is placed in a location that obstructs any straight line paths to the goal for the indicated trials. The obstacle is never placed directly over the goal position.
For 3D object pose estimation, we attached AprilTags~\cite{apriltag} to the top of the puzzle pieces, obstacles, and at the goal locations. An overhead Microsoft Azure Kinect sensor is used to retrieve images of the scene. A second Azure Kinect sensor records downward angled images of the scene as input to the autoencoder described in Section~\ref{sec:overall}. 
The Franka arm indirectly estimates end-effector forces and torques using the joint torques. Force detected relative to the end-effector is converted into a binary signal by setting a force threshold that detects contact with objects, but does not trigger during acceleration for movement in free space. 

\begin{figure}[t]
     \vspace{1em}
    \centering
    \includegraphics[width=0.4\textwidth]{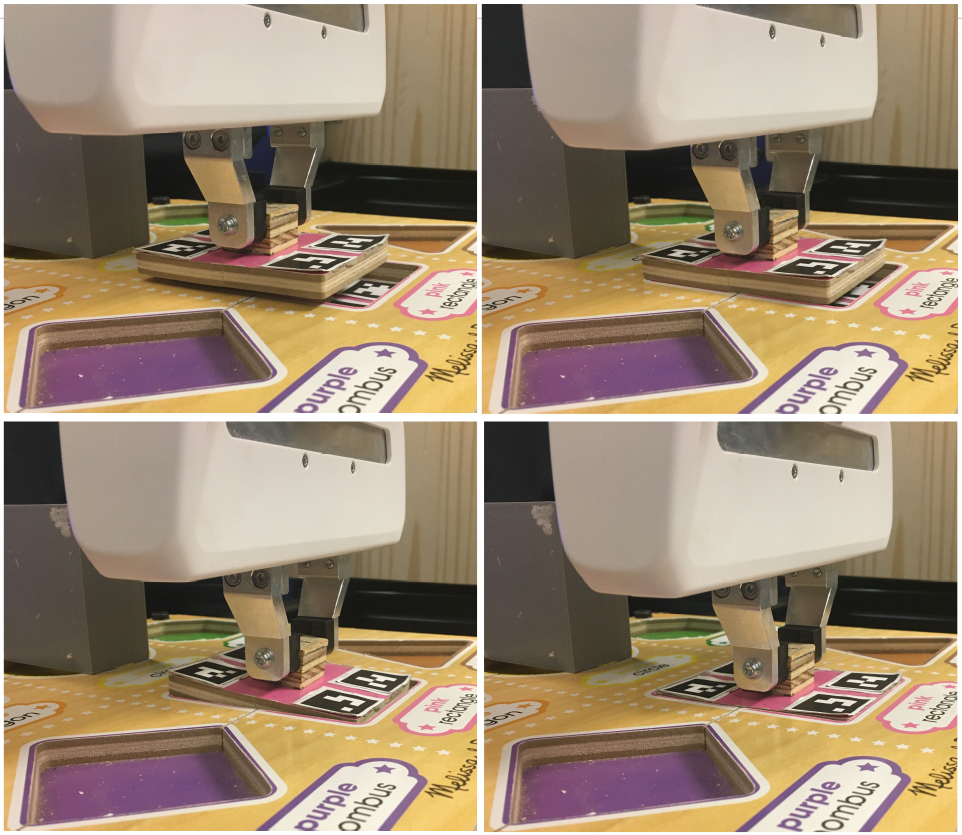}
    \vspace{-1em}
    \caption{Demonstration where the human operator moves the robot arm, via teleoperation, to complete the shape insertion task after model failure. We found that demonstrations worked best when the operator reset the shape to a position outside of the hole and performed a sliding motion into the hole.}
    \label{fig:demosetup}
    \vspace{-2em}
\end{figure}

\textbf{Baselines:} We compare our method to two different baselines. The first is the planner described in Section~\ref{sec:model-based-planner}. For the second baseline, we learned a policy using Imitation Learning (IL) in conjunction with Dynamic Movement Primitives (DMPs)~\cite{kroemer2016meta, schaaldmp}. For the DMP baseline, initial grasping is performed using the planner. The transport and insertion are done with DMPs. The DMP for this task was trained by collecting a trajectory using a kinesthetic human demonstration, while the obstacle was obstructing the path to the goal. More specifically, the DMP was trained from the oval starting position using the rectangle puzzle piece as the manipuland. Each translational Cartesian dimension was modeled as a separate DMP component. We used the formulation shown in Eq.~\ref{eq:dmp} and~\ref{eq:dmpforce}. We use the modification mentioned in Section IV-A of~\cite{kroemer2016meta} to include a goal state parameter that scales the DMP trajectory towards the goal state. This formulation also allows for the use of object features that can be used to scale the amplitude of a trajectory. This is useful if the start and end positions are close to one another, since the trajectory's amplitude can become sensitive to the goal state parameter's scaling. We scaled the $z$ dimension trajectory by 0.5 to quell this issue. The DMP parameters for execution are learned through linear ridge regression.
\vspace{-1.5em}

\begin{equation}
    \vec{y}=\alpha_{z}\left(\beta_{z} \tau^{-2}\left(y_{0}-y\right)-\tau^{-1} \dot{y}\right)+\tau^{-2} \sum_{j=1}^{M} \phi_{j} f\left(x ; \boldsymbol{w}_{j}\right)
    \label{eq:dmp}
\vspace{-1em}
\end{equation}

\begin{equation}
    f\left(x ; \boldsymbol{w}_{j}\right)=\alpha_{z} \beta_{z}\left(\frac{\sum_{k=1}^{K} \psi_{k}(x) w_{j k} x}{\sum_{k=1}^{K} \psi_{k}(x)}+w_{i 0} \psi_{0}(x)\right)
    \label{eq:dmpforce}
\vspace{-0.5em}
\end{equation}

\vspace{1em}
\begin{figure}[t]
    \centering
    \includegraphics[width=0.35\textwidth]{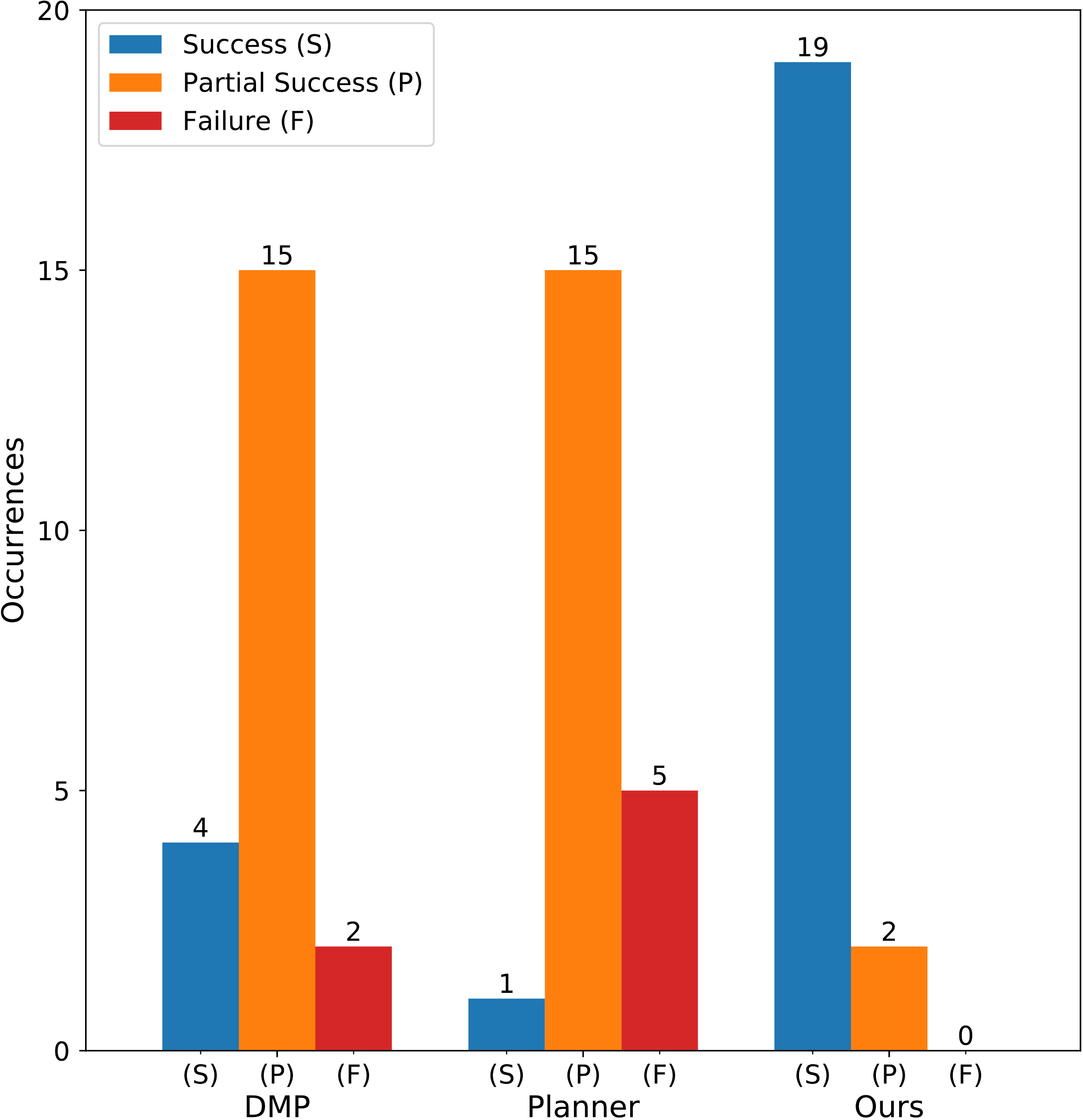}
    \vspace{-0.5em}
    \caption{Real world shape insertion results of all trials. The figure shows the number of times an outcome occurred for each implementation when there was no obstacle obstructing the path. Section~\ref{sec:ExperimentalResults} defines success.}
    \label{fig:successshape}
    \vspace{-2em}
\end{figure}

\subsection{Experimental Results}
\label{sec:ExperimentalResults}
\begin{figure}[t]
    \centering
    \includegraphics[width=0.35\textwidth]{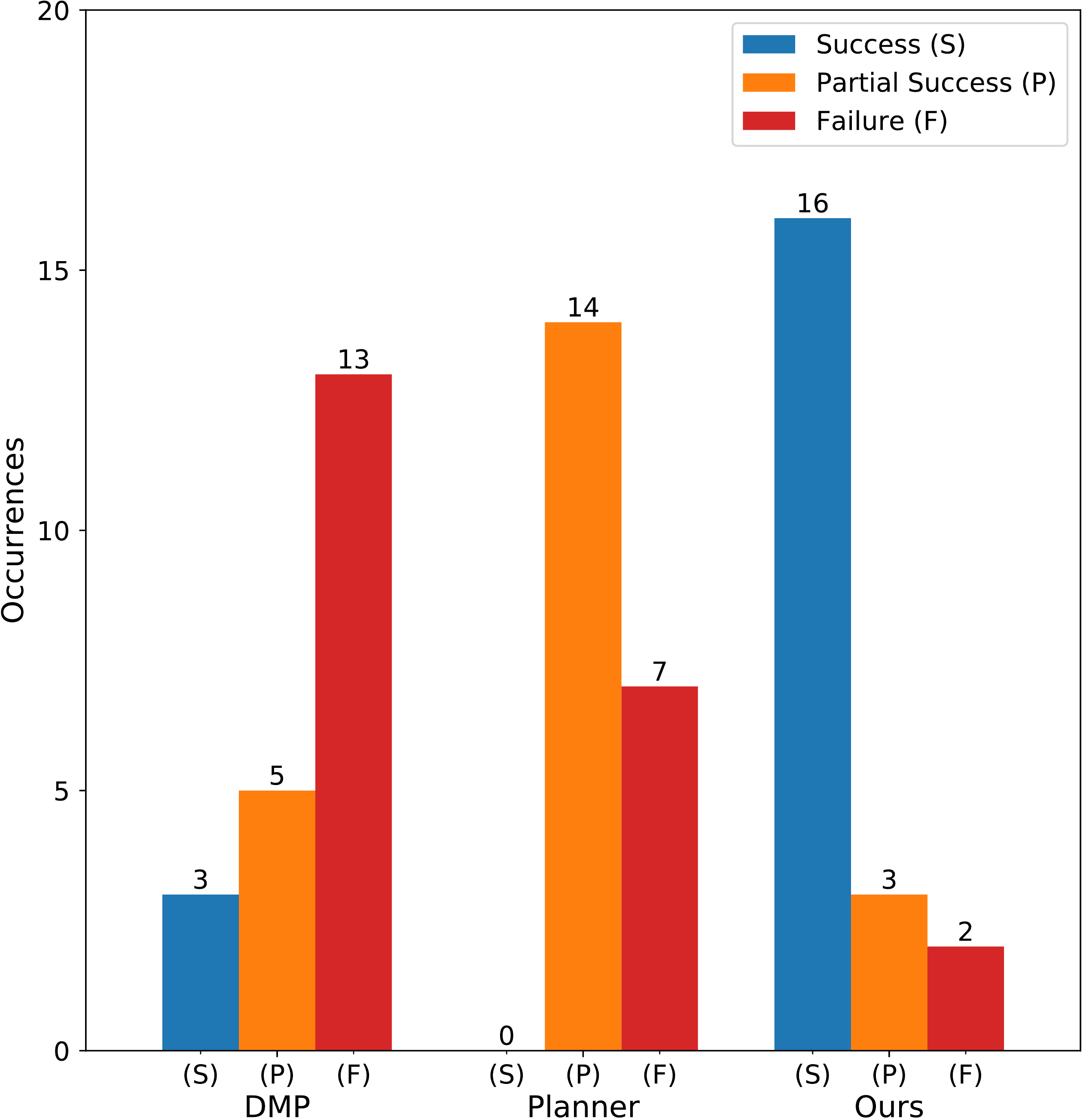}
    \vspace{-0.5em}
    \caption{Real world shape insertion results of all trials with an obstacle obstructing the path. The figure shows the number of times an outcome occurred for each implementation. Success is defined in Section~\ref{sec:ExperimentalResults}.}
    \vspace{-0.5em}
    \label{fig:obstacleshape}
\end{figure}

In this section, we show how our method compares to the baseline methods described in Section~\ref{sec:experimental} on two real-world tasks. The first task is to insert a puzzle piece into its corresponding hole. The experiments are performed with three shapes: a rectangle, a circle, and a square. Each shape goes into one of 8 corresponding slots in a $4\times2$ grid, shown in the bottom left of Figure~\ref{fig:pybullet}. We perform 7 trials for each shape where the starting position is one of the 8 regions, excluding the goal region. These experiments are then repeated with an obstacle obstructing the straight line path from start to goal. Success occurs when the shape is completely in the hole, see the bottom right of Figure~\ref{fig:demosetup}. A partial success is when only part of the shape is in the hole, see Figure~\ref{fig:failureexamples}. Failure indicates that either the robot hit an obstacle during execution or the shape was not in the hole at all. 
We show the performance of our method after training, as well as the baseline performances, in Figure~\ref{fig:successshape}. The results for the trials with an obstacle are shown in Figure~\ref{fig:obstacleshape}. 

\vspace{1em}

\begin{figure}[t]
    \centering
    \begin{tabular}{ c|c|c|c|c }
      \# Expert Demos: &1 & 5 & 10 & 20 \\
      \hline
      Success Rate: &$7/21$ & $15/21$ & $16/21$  &$19/21$ \\
    \end{tabular}
    \vspace{-0.5em}
    \caption{Overall success of our method on the shape insertion task depending on the number of training samples. The first row is the number of training samples used and the second row is the rate of success for the 21 trials. Success and the experimental trials performed are explained in~\ref{sec:ExperimentalResults}}.
    \label{fig:sampleeffect}
    \vspace{-1.5em}
\end{figure}

For the 21 shape insertion trials without the obstacle, we found that the method using the planner was only able to completely insert the puzzle pieces once, which occurred with the circle shape. The failure rate was $5/21$ and the partial success rate was $15/21$. The DMP baseline had a success rate of $4/21$, partial success rate of $15/21$, and failure rate of $5/21$. Lastly, our method had a success rate of $19/21$, partial success rate of $2/21$, and failure rate of $0/21$ when trained on 20 samples.

The next set of experiments show how well each method generalizes when an obstacle is introduced. The DMP method generates trajectories that collided with the obstacle $13/21$ times. Failures occur because the DMP parameters learned do not generalize over all possible obstacle configurations and starting locations. It should also be noted that the robot's joint configuration before the DMP was executed, i.e. the joint positions after the planner finished executing, seemed to affect the DMP's performance. We believe this was due to the goal configuration approaching the robot's joint limits from certain starting configurations. 

The method utilizing the planner hit the obstacle two times. This occurred when the task was constrained to a small area, causing the trajectory to go close to the obstacle and making the 1 cm perception error become more significant. An example of this is shown in Figure~\ref{fig:pybullet}. Our method, which relies on the planner for obstacle avoidance for most of the trajectory, also hits the obstacle $2/21$ trials. The remaining trials show performance similar to the trials without an obstacle, with a $16/21$ success and $3/21$ partial success rate. 
    
We also tested what effect using more trajectory demonstrations to train our method had on success rate. We observed an increase in success rate of insertion when more demonstrations were used. This is expected since the smaller $|\widetilde{\mathcal{D}}|$ is, the smaller the predicted region of $\tilde{S}$ becomes. Furthermore, more demonstrations lead to a better quality $\piMF$. These results are shown in Figure~\ref{fig:sampleeffect} and were performed without an obstacle.

\begin{figure}[t]
\centering
\vspace{1em}
\includegraphics[width=0.3\textwidth]{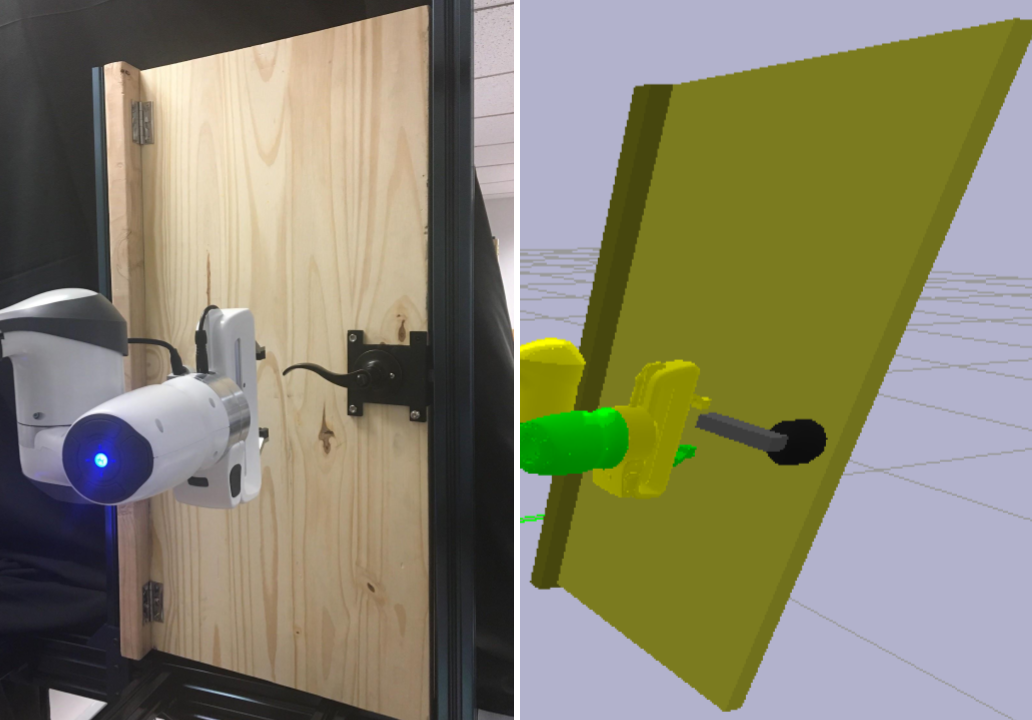}
\vspace{-0.7em}
\caption{Task setup for door opening. The real world environment is shown to the left and the simulation environment is shown to the right.}
\label{fig:doorenvsetup}
\vspace{-2em}
\end{figure}

The second task is a door opening task where the robot does not have access to an accurate model of the door handle. The environment we use for planning in simulation is shown in Fig~\ref{fig:doorenvsetup}. Note that the handle used in the planner is a simple rectangle, while the handle in our setup (shown in the left of~Fig~\ref{fig:doorenvsetup}) is more decorative and harder to grasp with parallel jaw grippers due to the curvature. 

We performed 10 trials of the task by executing actions from $\piPLAN$. All 10 trials led to successful opening of the door. There were no unexpected states observed during these trials, so $|\widetilde{\mathcal{D_{\mathcal{S}}}}| = 0$. Since the model-free method was not necessary, our framework only used the model-based planner and displayed the same results.
\vspace{-0.2em}

\vspace{-0.25em}
\section{Discussion}
\vspace{-0.25em}
\label{sec:discussion}
Model failure for the shape insertion task typically occurs shortly after non-sliding contact, e.g. contact with the target hole's border due to misalignment. Misalignment is mainly due to inaccurate localization of the hole from perception, which does not provide a pose of the hole that is accurate enough for this task. Another factor that affected alignment was the rotation that often occurred while puzzle pieces were being handled by the robot. Any deviation from the expected orientation can prevent the puzzle pieces from being inserted into the hole, since it can cause unexpected contact with the surrounding surface. We show examples of failures in Fig.~\ref{fig:failureexamples}. 

Once the model fails at time $t$, $s_{t}$ is then by definition in $\widetilde{S}$. This implies that the states around $s_{t}$ are also likely to be in $\widetilde{S}$, so we add $s_t$ to $\widetilde{S}$.
When we started collecting data at $s_{t}$, we found that in order for the expert to insert the peg, the shape needed to be moved out of the hole so it could be slid back in, as shown in Fig.~\ref{fig:demosetup}, which motivated our decision to have the agent change the goal state of the planner $S_{g}^*$ to be the initiation set of the skill $\mathcal{I}_{\textsc{SKILL}}$.

We observed that nearly all states in $\widetilde{\mathcal{S}}$ have low $z$-axis values, which corresponds to board contact or sliding. We also found that most of these states are clustered around the target location, which corresponds to the switch from $\piPLAN$ to $\piMF$ being close to the target location. A lower proportionality constant $k_0$ did lead to some spurious model failure detection. Measuring accumulated error rather than error between $s_t$ and $s_{t+1}$ could solve this.

Combining the planner with the learned skill allows for more of the task to be completed using the planner, which means a better overall system with a lower number of samples needed for training the local skill. For shape insertion, a reasonably reliable policy can be trained using only 5 data samples, although more samples were helpful. The trade-off for using only one data point with no domain-specific knowledge about symmetries is that the precondition space is smaller. For example, if the planner cannot find a plan to the precondition, such as if the states in $\mathcal{I}_{\textsc{SKILL}}$ were obstructed by obstacles, then $\piPLAN$ cannot be used to complete the task.

The door opening experiments demonstrated how our method can be used to identify which tasks need data to compensate for modelling deficiencies. In some tasks, the model is sufficient and there are no unexpected observations, meaning we can rely on the planner to generate trajectories and do not need expensive human demonstrations. 

This work is an example of a system that combines learning and planning algorithms to collect imitation trajectories only in the region of state space where it is necessary. We can leverage the ability of planners to generalize in regions where the model works and is fast enough to compute with, while also compensating for issues in perception error, modelling error, and search complexity. 

Improving the performance of sub-optimal planners using local learned skills has many potential directions for algorithmic development using local learned skills.
The primary limitation to our approach is that there is only one local $\piMF$. A more effective use of our method would be to learn multiple local policies or use more expressive policies (e.g. neural networks) with larger initiation sets. Additionally, integrating more sensory modalities, such as~\cite{lee2019making} did, would help mitigate issues caused by inaccuracies in perception.  


\vspace{-0.25em}
\section{Conclusion}
\vspace{-0.25em}
We proposed an approach that uses a planner, with a coarse probabilistic transition model, to find a trajectory and then switch to a model-free policy when the robot expects or observes model failure. During training, the robot learns a model of states from which we should learn a skill policy and collects expert demonstrations for the learned skill. We show results quantifying our method's improvement over pure planning or imitation learning for shape insertion. We also applied our method to a door opening task as another example of a contact-rich manipulation task. However, it did not need to learn a skill to complete the task. For future work, our method can be applied to dynamic tasks where approximate models are especially useful. Furthermore, we can train multiple local policies for different behaviours, select preconditions, and then choose between them through high-level planning. 
\vspace{-0.4em}
\vspace{-0.55em}
\section*{Acknowledgements}
\vspace{-0.55em}
We thank Kevin Zhang, Jacky Liang, Mohit Sharma, and many others for providing the infrastructure needed for the robot experiments. 

This work was in part supported by the Office of Naval Research under Grant No. N00014-18-1-2775, and the Army Research Laboratory under grant W911NF-18-2-0218 as part of the A2I2 program. Any opinions, findings, conclusions or recommendations expressed in this material are those of the author(s) and do not necessarily reflect the views of the ONR or ARL.

\end{document}